\documentclass[10pt,twocolumn,letterpaper]{article}

\usepackage[pagenumbers]{cvpr} % To force page numbers, e.g. for an arXiv version

\usepackage{graphicx}
\usepackage{amsmath}
\usepackage{amssymb}
\usepackage{epsfig}
\usepackage{graphicx}
\usepackage{booktabs}
\usepackage{textcomp}
\usepackage{subcaption}
\usepackage{algorithm}
\usepackage{booktabs}
\usepackage[noend]{algpseudocode}
\usepackage{tabularx}
\usepackage{tablefootnote}
\usepackage{listings}
\usepackage{multirow}
\usepackage{threeparttable}

\usepackage{xcolor}
\definecolor{lg}{RGB}{165,0,52}

\usepackage[accsupp]{axessibility}
\usepackage[pagebackref,breaklinks,colorlinks,allcolors=lg]{hyperref}

\usepackage[capitalize]{cleveref}
\crefname{section}{Sec.}{Secs.}
\Crefname{section}{Section}{Sections}
\Crefname{table}{Table}{Tables}
\crefname{table}{Tab.}{Tabs.}

\def\cvprPaperID{4567} % *** Enter the CVPR Paper ID here
\def\confName{CVPR}
\def\confYear{2022}

\begin{document}

%%%%%%%%% TITLE - PLEASE UPDATE
\title{L-Verse: Bidirectional Generation Between Image and Text}

\author{
Taehoon~Kim\thanks{Correspondence to: \texttt{taehoon.kim@lgresearch.ai}}\qquad Gwangmo~Song\qquad Sihaeng Lee  \qquad Sangyun~Kim\qquad Yewon~Seo \\
Soonyoung~Lee \qquad Seung~Hwan~Kim\qquad Honglak~Lee\qquad Kyunghoon~Bae \\
\\[-4mm]
LG AI Research
}
\maketitle

%%%%%%%%% ABSTRACT
\begin{abstract}
    Far beyond learning long-range interactions of natural language, transformers are becoming the de-facto standard for many vision tasks with their power and scalability. Especially with cross-modal tasks between image and text, vector quantized variational autoencoders (VQ-VAEs) are widely used to make a raw RGB image into a sequence of feature vectors. To better leverage the correlation between image and text, we propose L-Verse, a novel architecture consisting of feature-augmented variational autoencoder (AugVAE) and bidirectional auto-regressive transformer (BiART) for image-to-text and text-to-image generation. Our AugVAE shows the state-of-the-art reconstruction performance on ImageNet1K validation set, along with the robustness to unseen images in the wild. Unlike other models, BiART can distinguish between image (or text) as a conditional reference and a generation target. L-Verse can be directly used for image-to-text or text-to-image generation without any finetuning or extra object detection framework. In quantitative and qualitative experiments, L-Verse shows impressive results against previous methods in both image-to-text and text-to-image generation on MS-COCO Captions. We furthermore assess the scalability of L-Verse architecture on Conceptual Captions and present the initial result of bidirectional vision-language representation learning on general domain. 
\end{abstract}

%%%%%%%%% BODY TEXT
\section{Introduction}
\label{sec:intro}
Image-to-text and text-to-image generation and can be summarized as a task of learning cross-modal representations of image and text. Recent studies \cite{ramesh2021zeroshot, ding2021cogview, esser2021imagebart, Cornia_2020_CVPR} on vision-language tasks have highly improved the performance of each target task, in particular with various transformer architectures \cite{vaswani2017attention, Brown2020LanguageMA, devlin2019bert, child2019generating}. Initially designed to understand natural language, the \textit{dot-product multi-head attention mechanism} \cite{vaswani2017attention} effectively learns long-range interactions of sequential data.  
To leverage transformer architectures~\cite{vaswani2017attention} also in vision domains, an input image is factorized into a sequence of latent feature vectors. 

\begin{figure}[!tb]
  \centering
\includegraphics[width=\linewidth]{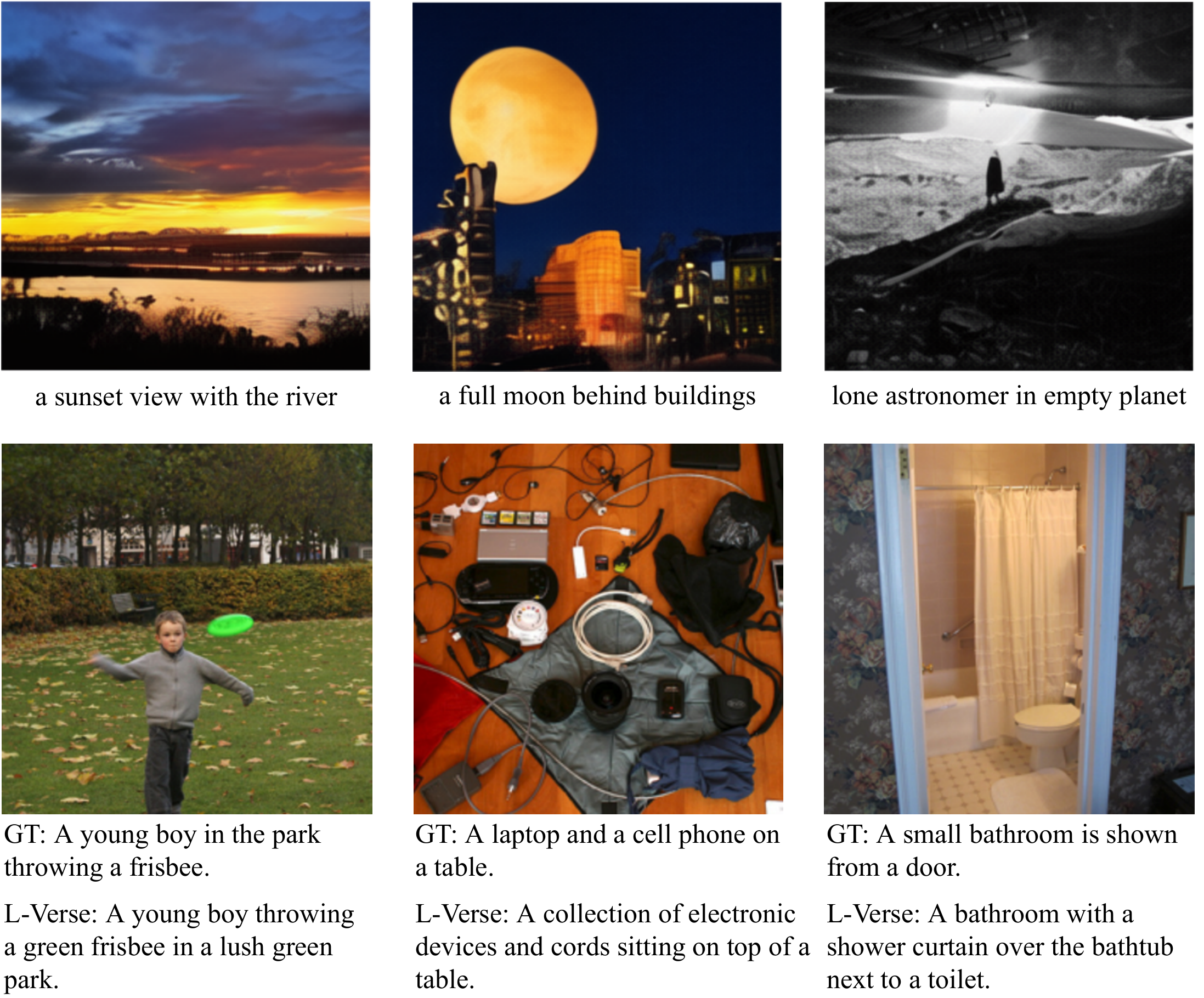}
\caption{
Examples of L-Verse on zero-shot text-to-image generation ($256 \times 256$ pixels) on Conceptual Captions \textit{(top)} and image-to-text generation on MS-COCO Captions \textit{(bottom)}. Trained in bidirectional manner, L-verse can both generate well-conditioned synthetic images and detailed captions without any finetuning. }
\label{fig1}
%\vspace{-4mm}
\end{figure}

To encode an image into a sequence of latent feature vectors, vector quantized variational autoencoder (VQ-VAE)~\cite{oord2018neural} can be used to learn a discrete latent representation with quantized embedding vectors from the \textit{visual codebook}.
VQ-VAE is a simple and powerful representation learning method to make image sequential and is widely used in conditional image generation tasks with auto-regressive pairs like RNNs \cite{oord2018neural,razavi2019generating} or transformers \cite{ramesh2021zeroshot, esser2021taming, esser2021imagebart, ding2021cogview}. Improving the reconstruction quality of VQ-VAE is also an active area of research \cite{ramesh2021zeroshot, esser2021taming,razavi2019generating}. 

Combining an auto-regressive  transformer \cite{Brown2020LanguageMA} with a feature extractor like VQ-VAEs or other deep convolutional neural networks (CNNs) is becoming a popular approach for various vision-language tasks. %
However, training a model for unidirectional image-to-text \cite{Cornia_2020_CVPR} or text-to-image \cite{ramesh2021zeroshot,ding2021cogview} generation task still requires a large amount of data or an extra object detection framework. We hypothesize that learning bidirectional cross-modal representation of image and text can alleviate this problem via better data efficiency.  

This paper proposes an approach, \textit{L-Verse (latent verse)}, for learning a bidirectional vision-language cross-modal representation. The key idea of L-Verse is two-fold: \textit{(i)} augment a visual codebook with diverse features and \textit{(ii)} enable an auto-regressive transformer to learn bidirectional image-text generation. 
Our novel \textit{cross-level feature augmentation technique} effectively increase the diversity of a visual codebook with unique feature embedding vectors. We furthermore add a \textit{segment embedding} to an auto-regressive transformer \cite{Brown2020LanguageMA} to teach the difference between image (or text) as given condition or generation target. 
Specifically, our contribution for vision-language cross-modal representation learning are summarized as follows:
\begin{itemize}
    \item We introduce a feature-augmented variational autoencoder (AugVAE), a VQ-VAE trained with cross-level feature augmentation. 
    With the feature-augmented visual codebook, AugVAE shows the state-of-the-art reconstruction performance on both in-domain ImageNet1K \cite{imagenet_cvpr09} validation set (Figure \ref{fig3}) and out-of-domain image datasets (Figure \ref{fig_trans}). 
    
    \item We propose a bidirectional auto-regressive transformer (BiART) for bidirectional image-text generation. We index each token with two different embedding vectors according to its role as a conditional reference (\texttt{[REF]}) or a generation target (\texttt{[GEN]}). With this segment embedding, our BiART can both generate corresponding images to given texts or meaningful captions to given images without any finetuning.
    
    \item  L-Verse, consisting of AugVAE and BiART, outperforms previously proposed image captioning models in most of the machine evaluation metrics on  MS-COCO Captions \cite{lin2015microsoft} \textit{Karpathy} test split. It is also notable that L-Verse does not require any object-detection framework, such as Faster-RCNN \cite{ren2015faster}.
    
     \item L-Verse shows comparable text-to-image generation results to other generative models on MS-COCO Captions \cite{lin2015microsoft}. We also assess the scalability of L-Verse for zero-shot text-to-image generation by training on Conceptual Captions \cite{sharma2018conceptual}.
\end{itemize}

 Section \ref{sec:related_work} briefly reviews previous works on VQ-VAE and cross-modal vision-language tasks.  Section \ref{sec:method} explains how we design AugVAE and BiART to learn the \textit{bidirectional} cross-modal representation between image and text. Section \ref{sec:experiments} shows quantitative and qualitative results on image reconstruction, image-to-text generation, and text-to-image generation. Section \ref{sec:conclusion} summarizes our paper with conclusion and discussion for future works.
 
%-------------------------------------------------------------------------
\section{Related Work}
\label{sec:related_work}
Adapting transformer architectures \cite{vaswani2017attention, Brown2020LanguageMA, devlin2019bert, child2019generating} for various vision-language tasks has been an active research area in the recent years. Since an image is a matrix of RGB pixel values, it should be first factorized into a sequence of feature vectors. Recent auto-regressive transformer based generative models \cite{ramesh2021zeroshot, ding2021cogview, esser2021taming} utilize different variants of VQ-VAE \cite{oord2018neural} to compress and reconstruct images. In this section, we introduce the main concept of VQ-VAE and its variants. We also explain how VQ-VAE or other CNN architectures are combined with auto-regressive transformers to solve image-to-text or text-to-image generation tasks.

\subsection{Vector Quantized Variational Autoencoder}
Vector quantized variational autoencoder, VQ-VAE \cite{oord2018neural}, is a set of an encoder $E$, a decoder $G$, and a visual codebook $Z$ for learning discrete representations of images. The CNN encoder $E$ factorizes the continuous representation of an image $\hat{z}$ into series of discrete vectors $z_q$, each selected from visual codebook $Z$. The CNN decoder $G$ is used to reconstruct any $z_q$ sampled from $Z$. Razavi \etal \cite{razavi2019generating} extend this approach to use hierarchical feature representation and apply exponential-moving-average (EMA) weight update to codebook $Z$. To better optimize the training of VQ-VAE,  Ramesh \etal \cite{ramesh2021zeroshot} use the gumbel-softmax relaxation \cite{jang2017categorical, maddison2017concrete}. Esser \etal \cite{esser2021taming} further improve the quality of image reconstruction with additional CNN discriminator, originated from generative adversarial network (GAN) \cite{NIPS2014_5423}.

\subsection{Image-to-Text Generation}
As the \textit{dot-product multi-head attention} \cite{vaswani2017attention} was initially designed for language tasks, transformers have achieved new state-of-the-art results in generating natural and detailed captions corresponding to an input image. Previous works \cite{Cornia_2020_CVPR, li2020oscar} utilize region features extracted using Faster R-CNN \cite{ren2015faster} to generate captions for each image. While visual semantics of each region improves the quality, objects outside detection target classes (80 classes for MS-COCO Detection \cite{lin2015microsoft}) get ignored.

\subsection{Text-to-Image Generation}
Generative adversarial networks (GANs) \cite{xu2017attngan, zhu2019dmgan,zhang2021crossmodal,tao2021dfgan} have been traditionally used for text-conditional image generation tasks. GAN based models focus on finding better modeling assumptions for specific data domains like CUB-200 \cite{WelinderEtal2010} or MS-COCO Captions \cite{lin2015microsoft}. Ramesh \etal \cite{ramesh2021zeroshot} first trained a 12-billion parameter transformer \cite{child2019generating} on 250-million image-text pairs for text-to-image generation in the general domain. Ding \etal \cite{ding2021cogview} proposed a 4-billion parameter transformer, CogView, with stable training techniques and finetuning strategies for various downstream tasks.

\begin{figure*}[!tb]
\vspace*{-0.05in}
  \centering
\includegraphics[width=\linewidth]{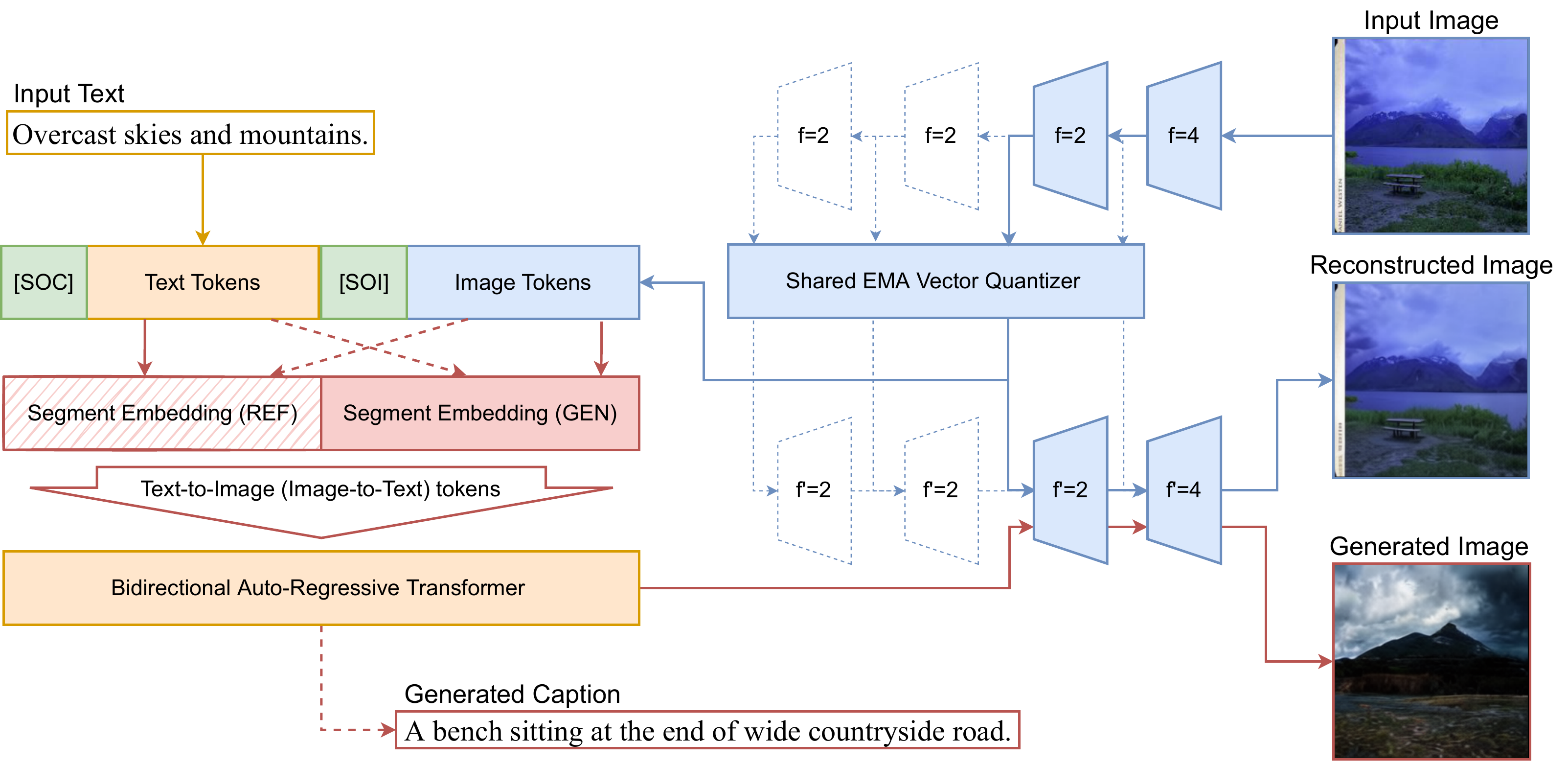}
\vspace*{-0.15in}
\caption{
Proposed L-Verse framework. \texttt{[SOC]}: Start of Caption (text) token. \texttt{[SOI]}: Start of Image token. Feature-augmented variational autoencoder (AugVAE) in \textit{blue}. Bidirectional auto-regressive transformer (BiART) in \textit{red}. AugVAE encoder $E$ encodes an image $x$ into tokens $z$. Segment embedding indicates each token as a conditional reference (\texttt{REF}), or a generation target (\texttt{GEN}). BiART $T$ either can generate image tokens $T(y)$ from text tokens $y$ or text tokens $T(z)$ from $z$. AugVAE decoder $G$ decodes $z$ and $T(y)$ into RGB images.}% Best viewed in color.}
\label{fig3}
\vspace{-4mm}
\end{figure*}
\section{Method}
\label{sec:method}
\subsection{Preliminary}
Ramesh \etal \cite{ramesh2021zeroshot} proposed a two-stage training procedure for text-to-image generation with an auto-regressive transformer \cite{Brown2020LanguageMA} : 
\begin{itemize}
\item \textbf{Stage 1:} Train a discrete variational autoencoder (dVAE) \cite{ramesh2021zeroshot} to compress each $256 \times 256$ RGB into a $32 \times 32$ grid of image tokens with each element of 8192 (\textit{$d_Z$}) possible values .

\item \textbf{Stage 2:} Concatenate up to 256 BPE-encoded text tokens with the $32 \times 32 = 1024$ image tokens, and train an auto-regressive transformer \cite{Brown2020LanguageMA} to model the joint distribution over text and image tokens.
\end{itemize}
The approach maximizes the evidence lower bound~\cite{kingma2014autoencoding, rezende2014stochastic} on the joint likelihood of the model distribution over the image $x$, the caption $y$, and the tokens $z$ for the encoded RGB image. From the factorization $p_{\theta,\psi}(x,y,z) = p_{\theta}(x|y,z)p_{\psi}(y,z)$, the lower bound is yielded as 
\begin{equation}
\begin{split}
    \ln{p_{\theta,\psi}(x,y)} \geq \mathop{\mathbb{E}}_{z \sim q_{\phi}(z|x)} (\ln{p_{\theta}(x|y,z)} \\
     -~D_{KL}(q_{\phi}(y,z|x), p_{\psi}(y,z)))
\end{split}
\end{equation}
where:
\begin{itemize}
    \item $q_{\phi}$ denotes the distribution over the $32 \times 32$ encoded tokens generated by dVAE encoder from the image $x$.
    \item $p_{\theta}$ denotes the distribution over the reconstructed image $\hat{x}$ from dVAE decoder.
    \item $p_{\psi}$ denotes the joint distribution over the text and image tokens modeled by the transformer.
\end{itemize}
In Stage 1, dVAE \textit{(or other VQ-VAE variants)} learns to minimize the reconstruction loss between $x$ and $\hat{x}$. In  Stage 2, an auto-regressive transformer optimizes two negative log-likelihood (NLL) losses: \textit{(i)} for caption $y$ and \textit{(ii)} for encoded image tokens $z$. 

\subsection{Proposed Approach: L-Verse Framework}
Inspired by DALL-E \cite{ramesh2021zeroshot}, we propose two major improvements for high-fidelity image reconstruction and bidirectional image-text generation:
\begin{itemize}
    \item We improve the diversity of a visual codebook $Z$ with cross-level feature augmentation. We first train multi-level (hierarchical) VQ-VAE (\textit{blue} in Figure \ref{fig3}) and apply weight-sharing to vector quantizers \cite{oord2018neural,razavi2019generating} in each feature-level. The hierarchical VQ-VAE is then finetuned to a VQ-VAE with codebook size $N = 32 \times 32$.
    \item We use segment embedding to indicate whether each token is given as a conditional reference (\texttt{[REF]}) or a generation target (\texttt{[GEN]}). For example, \texttt{[REF]} is added to each text token and \texttt{[GEN]} is added to each image token for text-to-image generation. %This simple idea enables both training and sampling in a bidirectional manner.  
\end{itemize}

Following subsections describe the training and sampling procedure of L-Verse in detail. The overview of L-Verse framework with actual reconstruction and generation examples are shown in Figure \ref{fig3}. 

\subsection{Feature-Augmented Variational Autoencoder}
\label{augvae}
\begin{figure}[!tb]
  \centering
\includegraphics[width=\linewidth]{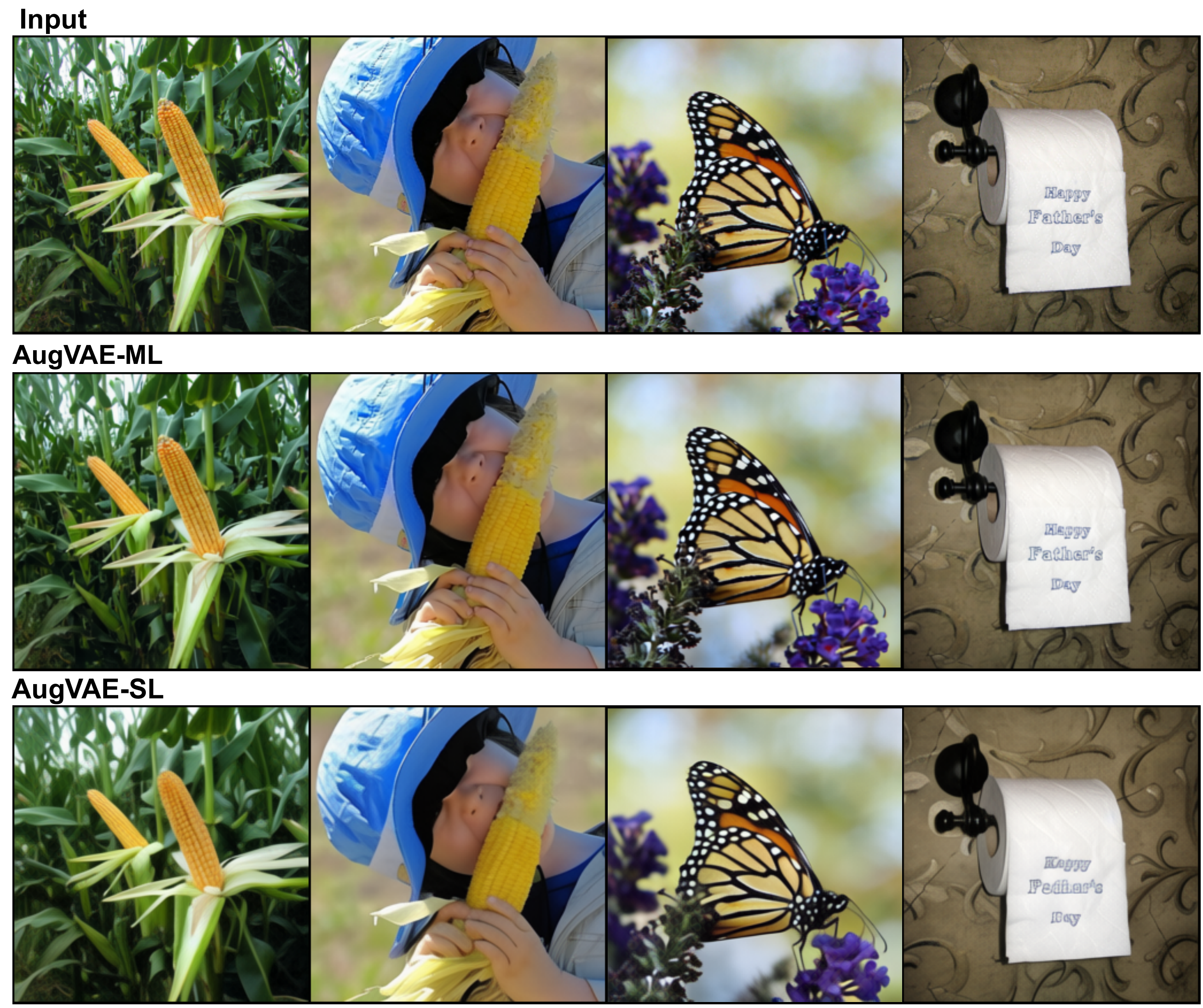}
\caption{Comparison of input images \textit{(top)}, 
reconstructions from multi-level (hierarchical) feature-augmented variational autoencoder (AugVAE-ML) \textit{(middle)}, 
and reconstructions from single-level feature-augmented variational autoencoder (AugVAE-SL) \textit{(bottom)} on Imagenet1K validation set. The resolution of each image is $256 \times 256$ pixels.
}
\label{fig2}
%\vspace{-4mm}
\end{figure}

Razavi \etal \cite{razavi2019generating} states that increasing the number of latent feature map adds extra details to the reconstruction. However, increasing the number of latent map also increases the total codebook size $N$, from $32 \times 32 = 1024$ \cite{oord2018neural} to $32 \times 32 + 64 \times 64 = 5120$ \cite{razavi2019generating}. 

\begin{figure}[!tb]
  \centering
\includegraphics[width=\linewidth]{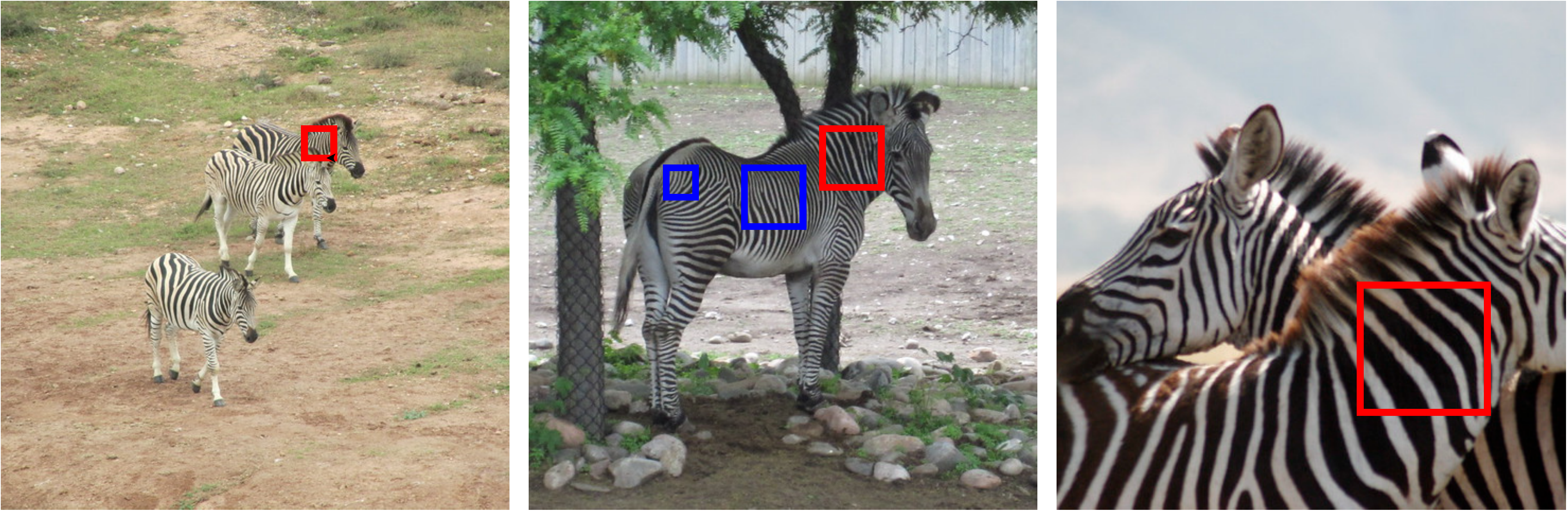}
\caption{
Cross-level patch similarity inside an image \textit{(blue)} and across images \textit{(red)}. Our feature-augmented variational autoencoder (AugVAE) utilizes the cross-level patch similarity to diversify the feature codebook.}
\label{fig4}
%\vspace{-4mm}
\end{figure}

For the high-quality image reconstruction at low-cost, we choose to use the single $32 \times 32$ latent map and augment the visual codebook $Z$ instead. From the example in Figure \ref{fig4}, similar patterns in various patch sizes can appear both in one image \textit{(blue)} and across different images \textit{(red)}. As the distance between similar patterns gets closer after vector quantization (VQ) \cite{oord2018neural}, extracting patches from different latent maps and storing them in one place removes duplicates and fills the codebook with unique 8192 ($d_Z$) possible values. 
% 8192 \textit{(dim $Z$)} possible values.

We optimize the encoder - vector quantizer - decoder architecture of VQ-VAE \cite{oord2018neural} for cross-level feature augmentation:
\begin{itemize}
    \item We define the encoder as $z = E(x, f,d_{out})$, where $x$ is an $n \times n \times d_{in}$ tensor and $f$ is a downsampling factor. $E(f, d_{out})$ downsamples a tensor $x$ into an $\frac{n}{f} \times \frac{n}{f} \times d_{out}$ tensor $z$. 
    
    \item We define the vector quantizer as $z_q = VQ(z, d_{Z})$, where $z$ is an $n \times n \times d$ tensor with continuous $d$-size vectors.  $z_q$ is a quantized version of $z$ with $d_{Z}$ possible values for each $d$-size feature vector. We use exponential-moving-average (EMA) vector quantizer \cite{razavi2019generating}. All vector quantizers in AugVAE shares weight parameters.
    
    \item We define the decoder as $\hat{x} = G(\hat{z}, f, d_{out})$, where $\hat{z}$ is an $n \times n \times d_{in}$ tensor and $f$ is a upsampling factor. $G(f, d_{out})$ upsamples an $n \times n \times d_{in}$ tensor $\hat{z}$ into an $nf \times nf \times d_{out}$ tensor $\hat{x}$. 
\end{itemize}

Hierarchical AugVAE (AugVAE-ML) consists of one $E(4, 256)$ and three $E(2, 256)$, four $VQ(8192)$ with shared weights, and three $G(2, 256)$ and one $G(4, 3)$. 
As shown in Figure \ref{fig3} with \textit{blue dotted and connected lines}, $E(4, 256)$ first downsamples a $256 \times 256 \times 3$ RGB image into a $64 \times 64 \times 256$ latent feature tensor. Each $E(2, 256)$ downsamples the previous tensor by 2. In total, four latent feature tensors ($64 \times 64 \times 256$, $32 \times 32 \times 256$, $16 \times 16 \times 256$, and $8 \times 8\times 256$) are extracted. These four tensors are quantized with a $VQ(8192)$ for each latent map. During the training of AugVAE-ML, each codebook with 8192 values gains diversity via weight sharing. Each $G(2, 256)$ upsamples the \texttt{concatenation} of previous tensor (if exists) and $\hat{z}$ of each level by 2. $G(4,3)$ reconstructs the original input from the last latent tensor and the quantized vector. 

To reduce the overall codebook size $N$, we finetune the AugVAE-ML into a single-level AugVAE (AugVAE-SL) of $32 \times 32$ latent map. We remove encoders and decoders with $16 \times 16$ and $8 \times 8$ latent map and replace the \texttt{concatenation} before each decoder with a \texttt{$1 \times 1$ convolution} to expand the last channel of previous latent tensor by 2. This modification to AugVAE-ML effectively stabilizes the finetuning process. The final architecture of AugVAE is depicted in Figure \ref{fig3} with \textit{blue connected lines}. As shown in Figure \ref{fig2}, AugVAEs can compress and reconstruct images with high-fidelity. Implementation details of AugVAE architecture and training hyperparameters are provided in Appendix \ref{app_augvae}. 

\subsection{Bidirectional Auto-Regressive Transformer}
With the \textit{masked dot-product multi-head attention}, the conventional auto-regressive transformer \cite{Brown2020LanguageMA} can only understand a given sequence from \textit{left to right}. Bidirectional generation between text and image doesn't require a transformer to be fully-bidirectional: learning how to distinguish an \texttt{image $\rightarrow$ text} sequence and a \texttt{text $\rightarrow$ image} sequence is enough.

We just tell our bidirectional auto-regressive transformer (BiART) whether the given text (or image) is a \textit{conditional reference} \texttt{([REF])} or a \textit{generation target} \texttt{([GEN])}. We feed BiART with an extra sequence of segment indexes for each token. A learnable embedding vector is assigned to each segment index \texttt{([REF])} and \texttt{([GEN])} and added to the input sequence. This simple idea enables the training and sampling of bidirectional image-text generation with BiART. 

For training, we feed the input sequence in \texttt{text $\rightarrow$ image} or \texttt{image $\rightarrow$ text} order alternately for each iteration. In each iteration, BiART optimizes two negative log-likelihood (NLL) losses:  \textit{(i)} for the conditional reference $y$ indexed as \texttt{[REF]} and \textit{(ii)} for the generation target $x$ indexed as \texttt{[GEN]}. When converges, BiART performs image-to-text (\textit{dotted red line in Figure \ref{fig3}}) and text-to-image (\textit{connected red line in Figure \ref{fig3}}) generations without any finetuning.

\subsection{Training Details}
\label{training_detail}
\begin{figure*}[!ht]
  \centering
\includegraphics[width=\linewidth]{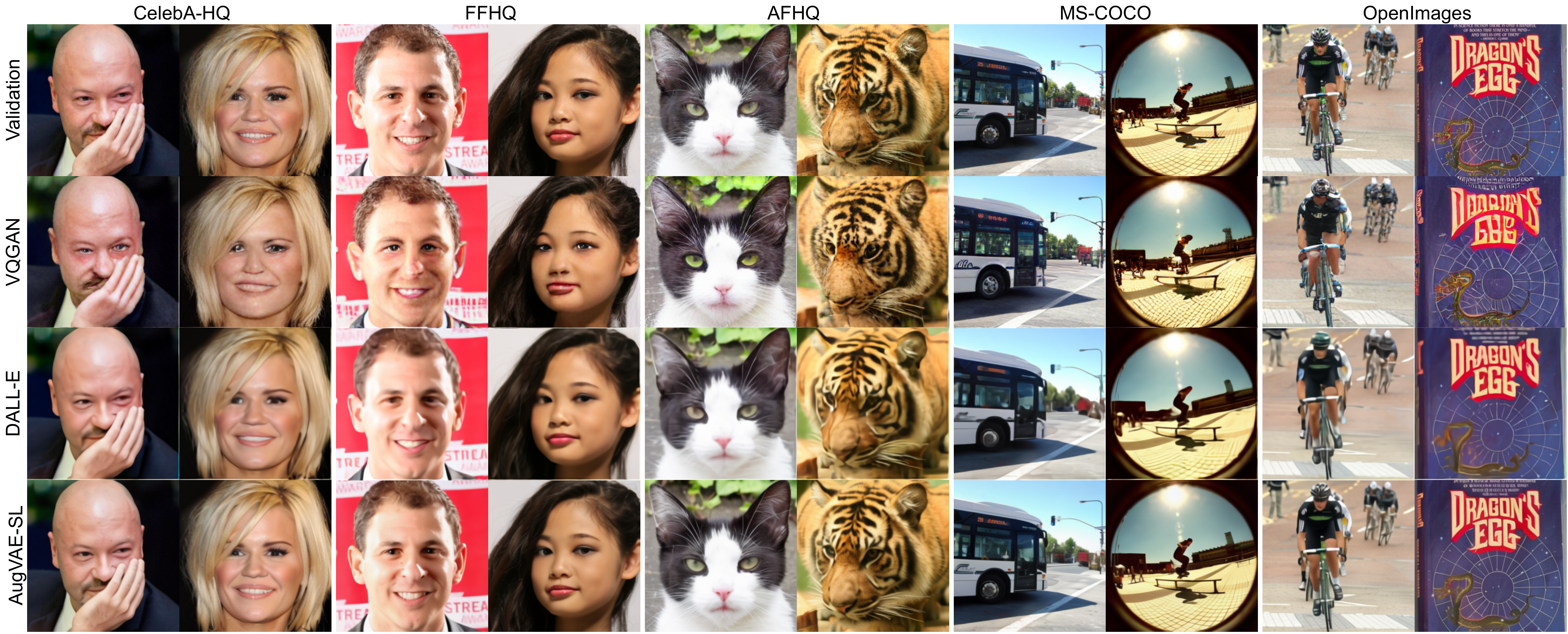}
\caption{
Qualitative evaluation on the reconstruction performance of different VQVAEs with unseen image domains. For all settings, we use ImageNet1K trained models without any finetuning. Images are resized to $256 \times 256$ with LANCZOS \cite{clark2018lanczos} filter. Cross-level feature augmentation allows AugVAE-SL to express out-of-domain unseen images in high-fidelity. Please zoom in for the detailed comparison.}% Best viewed in color.}
\label{fig_trans}
%\vspace{-4mm}
\end{figure*}

\paragraph{Architecture Overview} We first train 100-million parameter AugVAE-SL on ImageNet1K \cite{imagenet_cvpr09}. From results in Figure \ref{fig2}, \ref{fig_trans} and Table \ref{table:recon_fid}, our AugVAE-SL shows impressive reconstruction results with both in-domain and out-of-domain images. We use ImageNet1K-trained AugVAE-SL as encoder and decoder of L-Verse and pair encoded tokens with corresponding text tokens. BiART in L-Verse is 500-million parameter GPT \cite{Brown2020LanguageMA} transformer. While DALL-E \cite{ramesh2021zeroshot} and CogView \cite{ding2021cogview} use a sparse-transformer \cite{child2019generating} with custom attention masks for fast training and sampling, we use a GPT-style \cite{Brown2020LanguageMA} full-transformer to model the bidirectional cross-modal representation between image and text. We use 64 BPE-encoded \cite{sennrich2016neural} text tokens with 49408 possibilities and 1024 encoded image tokens with 8192 possibilities.
More details are provided in Appendix \ref{app_biart}.

\paragraph{Mixed Precision Training} To save computational cost and sampling time, BiART is trained with \texttt{FP16(O2)} mixed-precision training without inefficient stabilization methods like PB-relaxation \cite{ding2021cogview} or Sandwich-LayerNorm \cite{ding2021cogview}. These techniques are designed to eliminate the overflow in forward pass, but computationally inefficient. We instead inference AugVAE in FP32 to prevent the underflow caused by the vector quantizer. %Since we freeze AugVAE for Stage 2, only forward pass is performed. 

Ding \etal \cite{ding2021cogview} states that the precision problem in language-only training is not so significant as in text-to-image training. They hypothesize the heterogeneity of data as a cause. We found that training a transformer in bidirectional manner relieves the heterogeneity between image and text, and leads to stable training. In our toy experiments with smaller parameter sizes, BiART converged faster and showed better performance compared to previous image-to-text or text-to-image auto-regressive transformers. This states that bidirectional training approach with segment embedding is not only useful in the application-level, but also can be a new fundamental to find the cross-modal representation between different data domains.

\subsection{Sampling Details}
\label{text_sampling}
\paragraph{Image Sampling} Similar to Ramesh \etal \cite{ramesh2021zeroshot}, we rerank samples drawn from BiART using a pretrained contrastive model, CLIP \cite{radford2021learning}. CLIP assigns a score (\texttt{clip-score}) based on how well the image and text match with each other. For text-to-image generation, we make 64 samples from trained L-Verse model and calculate the \texttt{clip-score} to select a Top 1 image. We repeat this process $k$ times with different random seeds to sample $k$ images in total.

\paragraph{Text Sampling}
Our L-Verse auto-regressively generates a sequence of tokens. To generate an RGB image, 1024 ($32 \times 32$) tokens should be generated one-by-one. However, the length of text may vary depending on its reference image. For this reason, generating full 64 tokens doesn't always guarantee the quality of sampled text. In worst case, the result caption can be just a repeated sequence of same sentence and \texttt{[PAD]} tokens. From the statistics of MS-COCO Captions \cite{lin2015microsoft}, each caption contains average 16 words. We first sample 32 text tokens for each reference image and split the result caption by the full stop (\texttt{.}) token. We only use the first split to calculate the \texttt{clip-score} for reranking. This process dramatically saves computation time to generate 64 samples and select Top 1. 

From machine evaluation metrics in Table \ref{table:cap}, truncated captions from 32 tokens achieve new state-of-the-art in all metrics except CIDEr \cite{vedantam2015cider} among the peers trained only on MS-COCO Captions. L-Verse also shows comparable performance to OSCAR \cite{li2020oscar}, which is pretrained on 6.5-million image-text pairs. While full 64 token captions score 181.6 in CIDEr and 28.9 in SPICE \cite{anderson2016spice}, we figured out that scores are high just because each caption has more meaningful words. In our inner-group examination between full and truncated captions, we have agreed that each truncated version is more concise and accurate. We further investigate the quality of L-Verse generated captions with human evaluation, in comparison with human labeled ground-truths. 

\begin{table}[!htb]
\centering
\addtolength{\tabcolsep}{-1pt}
\begin{tabular}{lcccc}
\toprule
Model              & Codebook Size $N$       & $d_{Z}$ & FID & \\ 
\midrule
DALL-E \cite{ramesh2021zeroshot}            & $32 \times 32$             &  8192 & 32.01  & \\
VQGAN \cite{esser2021taming}             & $16 \times 16$             & 1024  & 7.94   &\\
VQGAN \cite{esser2021taming}              & $16 \times 16$              & 16384 & 4.98   &\\
\midrule
AugVAE-SL             & $32 \times 32$             & 8192  & \textbf{3.28}   &\\
\midrule
VQVAE-2 \cite{razavi2019generating}       & $64 \times 64$  \& $32 \times 32$  & 512   & $\sim$ 10  & \\
VQGAN  \cite{esser2021taming}             & $64 \times 64$ \& $32 \times 32$  &  512  & 1.45   &\\
\midrule

AugVAE-ML             & $64 \times 64 \sim 8 \times 8$  & 8192  & \textbf{1.04}   &\\
\midrule
\end{tabular}
\caption
{
Reconstruction Fréchet Inception Distance (FID) on ImageNet1K validation set. \textbf{$d_Z$}: The number of unique feature vectors in codebook. Both multi-level  (hierarchical) feature-augmented  variational autoencoder (AugVAE-ML) and single-level feature-augmented variational autoencoder (AugVAE-SL) achieve lowest FID among their peers.
}
\label{table:recon_fid}
\end{table}
\section{Experiments}
\label{sec:experiments}
In this section, we demonstrate the performance of proposed L-Verse in every aspect with both quantitative and qualitative experiments. We mainly discuss reconstruction performance on ImageNet1K \cite{imagenet_cvpr09} and out-of-domain unseen images, image-to-text generation (image captioning) results on MS-COCO Captions  \cite{lin2015microsoft}, text-to-image generation results on MS-COCO Captions. For MS-COCO, we trained L-Verse on MS-COCO Captions 2014 \textit{Karpathy} splits for fair evaluation with previous methods. We also include results of L-Verse trained on Conceptual Captions \cite{sharma2018conceptual} to further discuss the scalability of L-Verse architecture for zero-shot text-to-image generation. The FID can change depending on calculation tools. For fair comparison, we compute the Reconstruction FID with \texttt{torch-fidelity} \cite{obukhov2020torchfidelity}, caption evaluation metrics in Table \ref{table:cap} with \texttt{nlg-eval} \cite{sharma2017nlgeval}, and FIDs in Table \ref{table:gen_fid} with the \texttt{DM-GAN code} \cite{zhu2019dmgan}, available at \url{https://github.com/MinfengZhu/DM-GAN}. 

\subsection{Image Reconstruction}

As Esser \etal \cite{esser2021taming} stated, the reconstruction Fréchet Inception Distance (FID) \cite{Heusel2017GANsTB} of a VQ-VAE provide a lower bound on the achievable FID of the generative model trained on it. From the results on ImageNet1K validation set in Table \ref{table:recon_fid}, our AugVAE-ML trained with novel \textit{cross-level feature augmentation} achieves FID of \textbf{1.04}, meaning AugVAE-ML can compress and reconstruct image without nearly any information loss. Reconstruction examples on Figure \ref{fig2} also demonstrates AugVAE-ML's qualitative performance. Finetuned from AugVAE-ML, our AugVAE-SL also achieves new state-of-the-art FID of \textbf{3.28} among its single-level peers.

In a more difficult setting, we evaluate AugVAE-SL on reconstructing \emph{out-of-domain} unseen images.
From the examples in Figure \ref{fig_trans}, AugVAE-SL trained on ImageNet1K shows impressive reconstruction fidelity for all validation input images without extra finetuning. From this result,we believe that our AugVAE-SL can work as a new \textit{``imagenet-backbone"} for various vision tasks. Detailed examination with more examples for each dataset in Figure \ref{fig_trans} can be found in Appendix \ref{app_recon}.

\begin{table}[!htb]
\centering
\addtolength{\tabcolsep}{-2pt}
\begin{threeparttable}[t]
\begin{tabular}{lcccccc}
\toprule
Model              & B-4   & M    & R     & C     & S   & \\ 
\midrule
SCST \cite{rennie2017selfcritical}              & 34.2  & 26.7  & 55.7  & 114.0 &  -   &\\
Up-Down \cite{anderson2018bottomup}            & 36.3  & 27.7  & 56.9  & 120.1 & 21.4 &\\
RFNet \cite{shen2019rfnet}             & 36.5  & 27.7  & 57.3  & 121.9  & 21.2 &\\
Up-Down+HIP \cite{yao2019hierarchy}        & 38.2  & 28.4  & 58.3  & 127.2  & 21.9 &\\
GCN-LSTM \cite{marcheggiani2017encoding}          & 38.2  & 28.5  & 58.3  & 127.6  & 22.0 &\\
SGAE \cite{yang2018autoencoding}               & 38.4  & 28.4  & 58.6  & 127.8  & 22.1 &\\
ORT \cite{herdade2020image}              & 38.6  & 28.7  & 58.4  & 128.3  & 22.6 &\\
AOANet \cite{huang2019attention}             & 38.9  & 29.2  & 58.8  & 129.8 & 22.4 &\\
$M^2$ Transformer \cite{Cornia_2020_CVPR}  & 39.1  & 29.2  & 58.6  & 131.2 & 22.6 &\\

\midrule
L-verse          & \textbf{39.9}  & \textbf{31.4}  & \textbf{60.4} & 102.2 & \textbf{23.3} &\\
\tnote{*}  L-verse          & 27.6  & 23.6  & 43.9 & \textbf{181.6} & \textbf{28.9}  &\\
\midrule
\tnote{$\dagger$} $OSCAR_B$ \cite{li2020oscar}             & 40.5  & 29.7  &  -    & 137.6 & 22.8 &\\
\tnote{$\dagger$} $OSCAR_L$ \cite{li2020oscar}          & 41.7  & 30.6  &  -    & 140.0 & 24.5 &\\
\midrule
\end{tabular}
\begin{tablenotes}
   \item[-] \footnotesize{\textbf{B-4}: BLEU-4  \textbf{M}: METEOR  \textbf{R}: ROUGE \textbf{C}: CIDEr  \textbf{S}: SPICE}
   \item[*] \footnotesize{Captions generated without truncation.}
   \item[$\dagger$] \footnotesize{Models pretrained on 6.5 million image-text pairs.}
\end{tablenotes}

\end{threeparttable}
\caption
{
Comparison with state-of-the-arts on MS-COCO Captions \textit{Karpathy} test split. We mainly compare results with models trained only on MS-COCO. Results from OSCAR (which requires additional fine-tuning) is given as a reference.
}
\label{table:cap}
\end{table}

\subsection{Image-to-Text Generation}
\label{image_text}
We evaluate the image-to-text generation (image captioning) performance of L-Verse with \textit{(i)} machine evaluation metrics against previous MS-COCO trained state-of-the-arts and \textit{(ii)} human evaluation against corresponding ground-truth (reference) captions.

\paragraph{Machine Evaluation}
We first compare the performance of our model with MS-COCO trained image captioning models in Table \ref{table:cap}. We also include OSCAR \cite{li2020oscar}, which is finetuned from a pretrained model with 6.5-million image-text pairs, to assess the scalability of our model with larger dataset. With proposed sampling method in Section \ref{text_sampling}, L-Verse surpasses all the other methods in terms of BLEU-4, METEOR, ROUGE, and SPICE without any object detection framework or other extra information. L-Verse also shows comparable performance to OSCAR, showing that pretraining L-Verse on a larger set of image-text pairs is a promising direction for future work. 

\begin{figure}[!tb]
  \centering
\includegraphics[width=\linewidth]{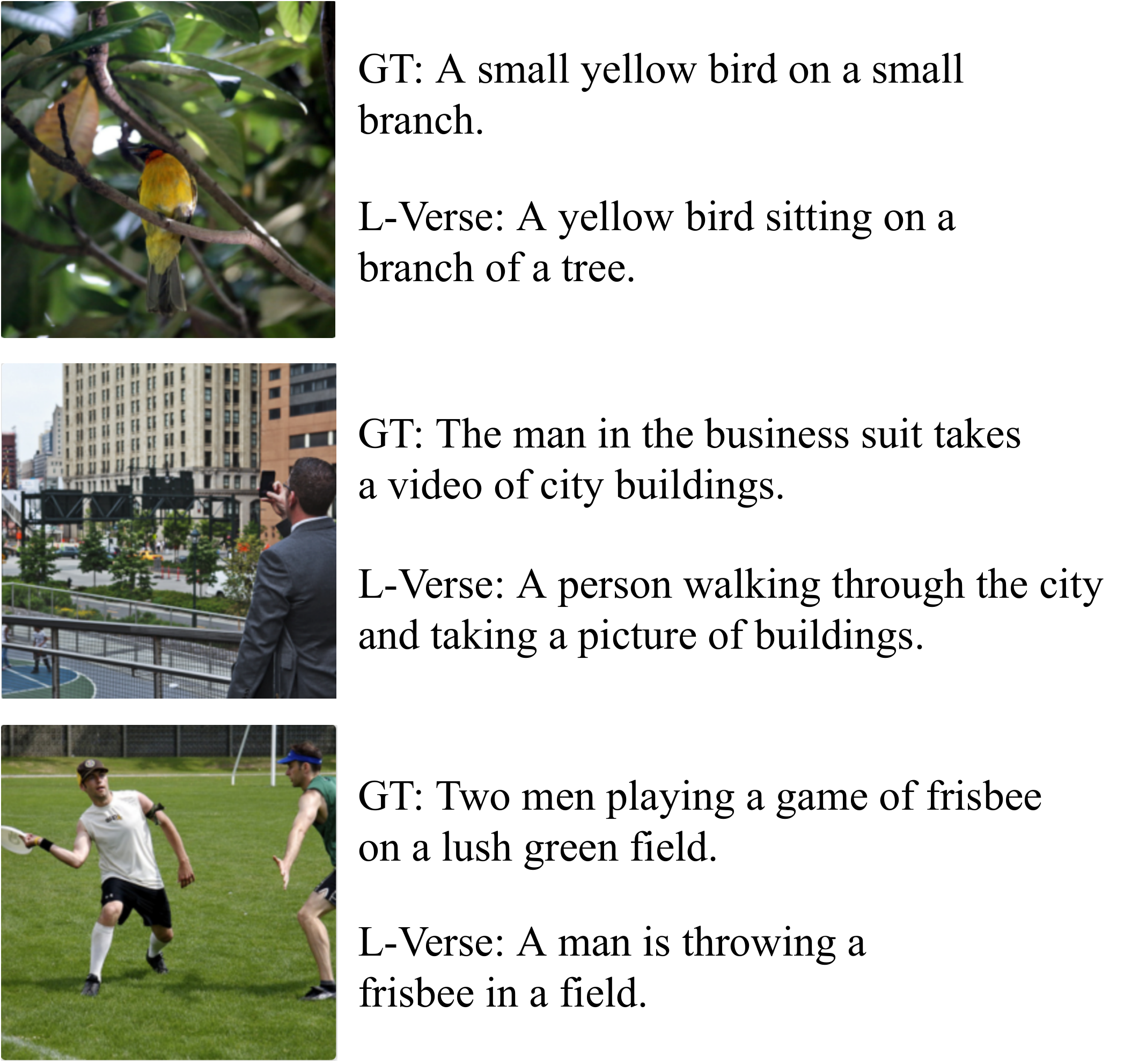}
\caption{ Examples of captions generated by L-Verse with corresponding ground-truths. Examples are sampled from conducted human evaluation results which received \textit{``Both captions well describe the image"}. }
\label{fig_cap}
%\vspace{-4mm}
\end{figure}

\paragraph{Human Evaluation}
Without caption truncation, L-Verse achieves the highest score in CIDEr and SPICE. As we stated in Section \ref{text_sampling}, machine evaluation metrics don't always guarantee the qualitative performance of generated captions. We further conduct a human evaluation similar to the one used in Li \etal \cite{li2016deep}. We directly evaluate L-Verse generated captions with human-labeled ground-truth captions, which is the theoretical upper-bound of L-Verse in image-to-text generation. We randomly sample 500 sets of images, corresponding ground-truth caption (GT), and L-Verse generated caption (Pred) from MS-COCO 2014 \textit{minival} split for the evaluation pool. 150 anonymous people participated for the evaluation. For each participant, we show randomly sampled 50 sets of image, GT, and Pred from the pool and ask to choose the best caption for each set. To cope with \textbf{tie} situation, we also allow each participant to choose \textit{``Both captions well describe the image"}. We provide more details on human evaluation in Appendix \ref{app_i2t}.
Results in Figure \ref{fig_human} show that L-Verse can generate a detailed explanation of a given image, receiving \textbf{30.4\%} of votes (Pred + Both) in total. Examples in Figure \ref{fig_cap} also demonstrate that L-Verse doesn't miss the detail of each image.

\begin{figure}[!tb]
  \centering
\includegraphics[width=\linewidth]{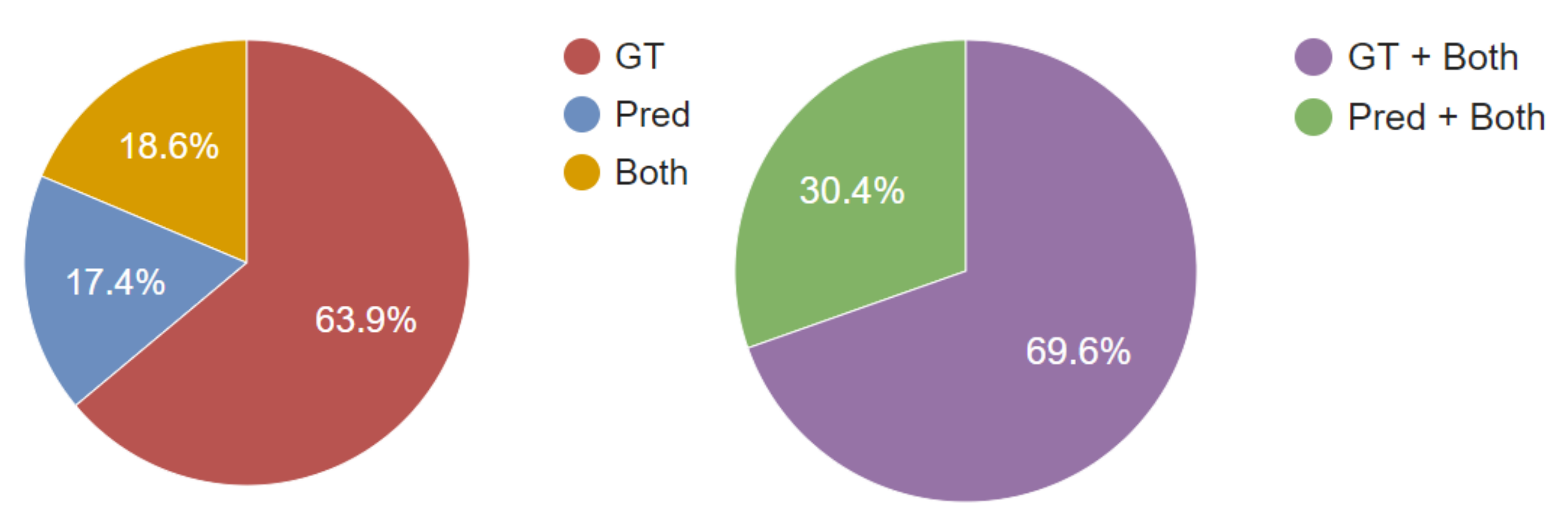}
\caption{
Human evaluation results on MS-COCO Captions \textit{minival} split. With question \textit{``Which caption well describes the given image?"}, L-Verse generated captions received \textbf{30.4\%} of votes (Pred + Both) in total.}
\label{fig_human}
\vspace{-2mm}
%\vspace{-4mm}
\end{figure}

\vspace*{-2mm}
\subsection{Text-to-Image Generation}
\begin{table}[!tb]
\centering
\addtolength{\tabcolsep}{-2.5pt}
\begin{threeparttable}[t]
\begin{tabular}{lcccccc}
\toprule
Model               & FID-0         & FID-1 & FID-2 & FID-4 &  FID-8 & \\ 
\midrule
AttnGAN \cite{xu2017attngan}            & 35.2           & 44.0           & 72.0          & 108.0         & 100.0 & \\
DM-GAN \cite{zhu2019dmgan}             & \textbf{26.0}  & 39.0           & 73.0          & 119.0         & 112.3 & \\
DF-GAN \cite{tao2021dfgan}             & \textbf{26.0}  & \textbf{33.8}  & 55.9          & 91.0          & 97.0  &\\
\midrule
L-Verse                                & 45.8           &  41.9         & \textbf{35.5} & \textbf{30.2} &  \textbf{29.8} &\\
\midrule
\midrule
%\toprule
\tnote{*} L-Verse-CC            & 37.2          &  31.6          &  25.7         & 21.4        &  \textbf{21.1} & \\
\midrule
\tnote{$\dagger$} DALL-E \cite{ramesh2021zeroshot}             & 27.5           & 28.0           & 45.5          & 83.5          & 85.0  & \\
\tnote{$\dagger$} CogView \cite{ding2021cogview}            & 27.1           & 19.4           & 13.9          & 19.4          & 23.6  & \\

\midrule
\end{tabular}
\begin{tablenotes}
   \item [-] \footnotesize{\textbf{FID-$k$}: FID of images blurred by radius $k$ Gaussian filter.}
   \item[*] \footnotesize{L-Verse trained on Conceptual Captions. }    
   \item[$\dagger$] \footnotesize{Models trained on over 30 million image-text pairs.}
\end{tablenotes}

\end{threeparttable}
\vspace*{-1mm}
\caption
{
Fréchet Inception Distance (FID) on a subset of 30,000 captions sampled from MS-COCO Captions validation set. We mainly compare results with models trained only on MS-COCO. In the bottom part of the table, we provide results from DALL-E, Cogview, and L-Verse-CC (which are trained from much larger datasets) as references.
}
\label{table:gen_fid}
\vspace*{-3mm}
\end{table}

\vspace*{-1mm}
Following Ramesh \etal \cite{ramesh2021zeroshot} and Ding \etal \cite{ding2021cogview}, we evaluate the text-to-image generation performance of L-Verse by comparing it to prior approaches. We compute FIDs in Table \ref{table:gen_fid} after applying a Gaussian filter with varying radii to both validation images and samples from L-Verse. We use the image sampling process explained in Section \ref{text_sampling}. Generated samples with corresponding captions from MS-COCO are provided in Appendix \ref{app_t2i}.

According to Ramesh \etal \cite{ramesh2021zeroshot}, training a transformer on tokens from a VQ-VAE encoder disadvantages model since it generates an image in low-frequency domain. Trained on same MS-COCO training set, L-Verse achieves best FID among previous approaches by a large margin with a slight blur of radius 2. The gap tends to increase as the blur radius is increased. We also compare L-Verse-CC, L-Verse trained on Conceptual Captions \cite{sharma2018conceptual}, with DALL-E \cite{ramesh2021zeroshot} and CogView \cite{ding2021cogview}. Considering the size of training data, L-Verse shows comparable text-to-image generation performance to other large-scale transformers as blur radius increases. 

\begin{figure}[!tb]
  \centering
\includegraphics[width=\linewidth]{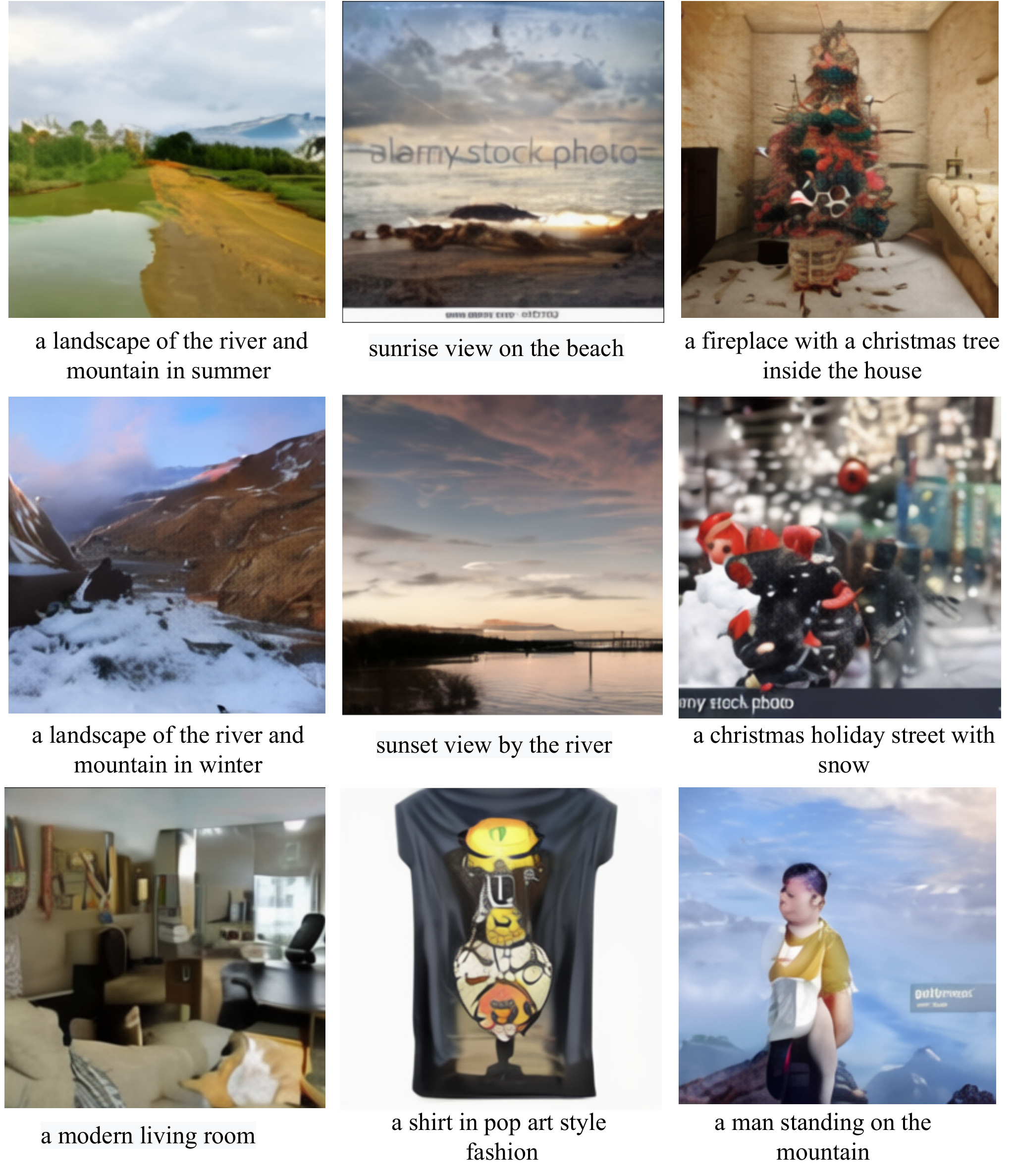}
\caption{
Examples of zero-shot text-to-image generation. Results are sampled from L-Verse-CC, which is trained on 3-million image-text pairs from Conceptual Captions. The resolution of each image is $256 \times 256$ pixels.}
\label{fig_coco}
\vspace*{-4mm}
%\vspace{-4mm}
\end{figure}

It is interesting that L-Verse shows decreasing FID with increasing blur radius, while other models show increasing FID. We hypothesize that L-Verse focuses on objects in the reference text, showing lower FIDs when high-frequency details are lost. This finding also corresponds with image-to-text generation results in Section \ref{image_text}. We also provide initial zero-shot text-to-image generation results with L-Verse in Figure \ref{fig_coco}. Trained on Conceptual Captions \cite{sharma2018conceptual}, L-Verse generates detailed images with objects in reference texts. We believe that L-Verse will also be able to generate realistic images in zero-shot fashion when trained with sufficient data and scale.

\section{Conclusion}
\label{sec:conclusion}
This paper presents L-Verse, a novel framework for bidirectional generation between image and text. Our \textit{feature-augmented variational autoencoder} (AugVAE) achieves new state-of-the-art reconstruction FID and shows its potential as an universal backbone encoder-decoder for generative models. We also enable bidirectional training of auto-regressive transformer with \textit{segment embedding}. Proposed \textit{bidirectional auto-regressive transformer} (BiART) learns both image-to-text and text-to-image as a whole. Experimental results demonstrate that our L-Verse framework shows remarkable performance in both image-to-text and text-to-image generation. 

\section*{Acknowledgments}
We would first like to thank Yountae Jung and LG AI Research AI Platform Sector for providing technical support and computing resources for training and evaluation. We would also like to thank Jinsu Yoo, Daechul Ahn, Janghyeon Lee, Yeonsik Jo, and other members of LG AI Research Vision Lab for helpful discussions and feedback on the paper. Finally, we would like to thank all members of LG AI Research, without whom this work would not have been possible.

%%%%%%%%% REFERENCES
{\small
\bibliographystyle{ieee_fullname}
\bibliography{egbib}
}

\clearpage
\appendix
\section{Details for AugVAE}
\label{app_augvae}
\subsection{Architecture}
\begin{figure*}[!tb]
  \centering
\includegraphics[width=0.9\linewidth]{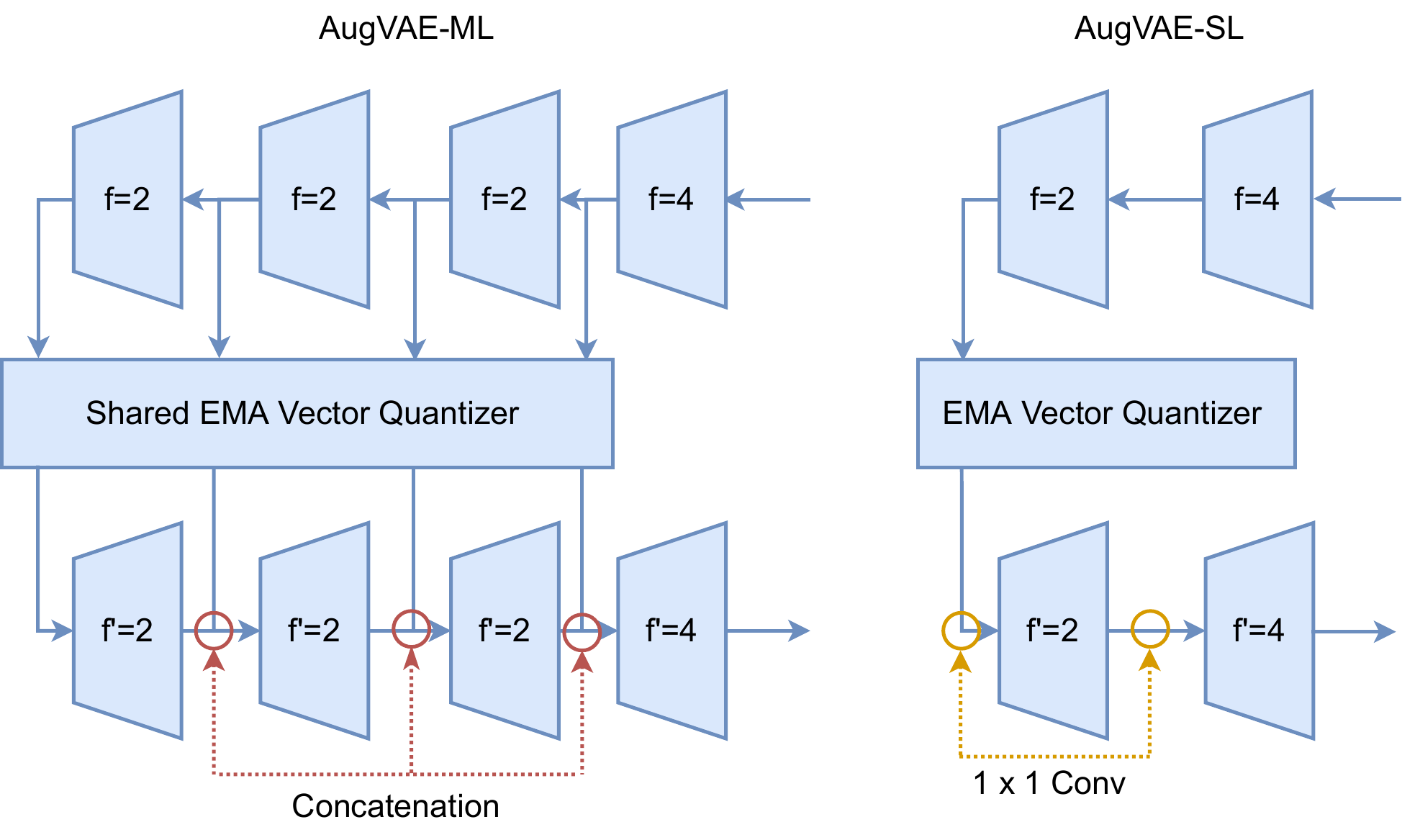}
\caption{
Proposed AugVAE. Trained with cross-level feature augmentation, AugVAE-ML is finetuned into AugVAE-SL to reduce the length of encoded image sequence. We remove unnecessary encoders and decoders from AugVAE-ML and replace the concatenation operation with a $1 \times 1$ convolution which expands the last dimension of the input tensor by two.}
\label{fig_aug_detailed}
\end{figure*}

The AugVAE encoder and decoder are ResNet \cite{he2016deep} with bottleneck-style Resblocks. Our AugVAE is specifically based on the encoder-decoder from official VQGAN \cite{esser2021taming} implementation available at \url{https://github.com/CompVis/taming-transformers}. From VQGAN implementation, we removed the attention block and applied the modification we describe in \ref{augvae}.  The high-level architecture of our AugVAE is depicted in Figure \ref{fig_aug_detailed}. Before we start AugVAE-SL fine-tuning, we change the model architecture by removing $16\times16$ and $8\times8$ latent map from AugVAE-ML and replacing \texttt{concatenation} with \texttt{$1\times1$ convolution} for channel upsampling. Precise details for the architecture are given in files \texttt{latent-verse/models/vqvae.py} and \texttt{latent-verse/modules/vqvae/vae.py} of our source code available at: \url{https://github.com/tgisaturday/L-Verse}.

\subsection{Training}
Our AugVAE is trained on ImageNet1K \cite{imagenet_cvpr09}. We resize each image into $256 \times 256 \times 3$ and apply random crop with 0.75 crop ratio for training. We train both AugVAE-ML and AugVAE-SL using AdamW \cite{DBLP:journals/corr/abs-1711-05101} optimizer with $\beta_1 = 0.9$, $\beta_2 = 0.999$, $\epsilon = 10e-8$, weight decay multiplier $1e-5$, and the learning rate $4.5e-6$ multiplied by the batch size. We half the learning rate each time the training loss appeared to plateau. For the loss term, we use a combination of mean-squared-error (MSE) and LPIPS \cite{zhang2018unreasonable} losses between the input and the reconstructed image. For stable training, we multiply the LPIPS loss by 0.1.

\section{Details for BiART}
\label{app_biart}
\subsection{Architecture}
Our BiART is similar to the GPT architecture \cite{Brown2020LanguageMA}. We utilize the \texttt{minGPT} implementation of GPT architectures available at \url{https://github.com/karpathy/minGPT}. We only add \texttt{segment embedding} with dimension size 256 for \texttt{[REF]} and \texttt{[GEN]}. Each \texttt{segment embedding} is added to the \texttt{positional encoding} of an input token. We use a 32-layer decoder-only transformer with 1024 dimensional states and 16 masked self-attention heads. While BiART uses an integrated embedding matrix for image and text tokens, each token groups are separately indexed from 0 to 8191 and from 8192 to 57599. Special tokens \texttt{[PAD]} (\textit{padding}), \texttt{[SOC]} (\textit{start-of-text}), and \texttt{[SOI]} (\textit{start-of-image}) are indexed from 57600 to 57602.

\subsection{Training}
BiART is trained on MS-COCO Captions \cite{lin2015microsoft} and Conceptual Captions \cite{sharma2018conceptual}. We resize each image into $256 \times 256 \times 3$ and apply random crop with 0.75 crop ratio for training. We apply BPE dropout \cite{provilkov-etal-2020-bpe} with a rate of 0.1 to our byte-pair encoder. We also apply residual, embedding, and attention dropouts \cite{JMLR:v15:srivastava14a} with a rate of 0.1.  We train BiART using AdamW \cite{DBLP:journals/corr/abs-1711-05101} optimizer with $\beta_1 = 0.9$, $\beta_2 = 0.95$, $\epsilon = 1e-8$, weight decay multiplier $1e-2$, and the learning rate $4.5e-7$ multiplied by the batch size. We don't apply weight decay to embedding parameters. We half the learning rate each time the training loss appeared to plateau.

\section{Examples for Image Reconstruction}
\label{app_recon}
We provide more examples of in-domain image reconstruction in Figure \ref{in_domain} and out-of domain in Figure \ref{out_domain}. We also provide the reconstruction FID of AugVAE-SL on various datasets in Table \ref{table:trans_fid} as a reference for future works. AugVAE-SL trained on ImageNet1K shows ``$\le 8$" FID for all data domain without extra finetuning. The resolution of each reconstructed image is $256\times256$.
\begin{table}[!tb]
\centering
\begin{threeparttable}[t]
\begin{tabular}{lcc}
\toprule
Dataset              &  FID & \\ 
\midrule
CelebA-HQ \cite{liu2015faceattributes}          &   7.24       &  \\
FFHQ \cite{karras2019stylebased}             & 4.92 & \\
AFHQ \cite{choi2020starganv2}        & 4.36 & \\
MS-COCO \cite{lin2015microsoft}             & 4.77 &\\
OpenImages V6 \cite{2020openimages}          & 3.15  &\\

\midrule
\end{tabular}
\end{threeparttable}
\caption
{
Reconstruction Fréchet Inception Distance (FID) of AugVAE on various datasets. For all settings, we use ImageNet1K trained AugVAE-SL without any finetuning on each dataset. Images are resized to $256 \times 256$ with LANCZOS \cite{clark2018lanczos} filter. 
}
\label{table:trans_fid}
\end{table}

\section{Examples for Image-to-Text Generation}
\label{app_i2t}
\begin{figure*}[!tb]
  \centering
\includegraphics[width=\linewidth]{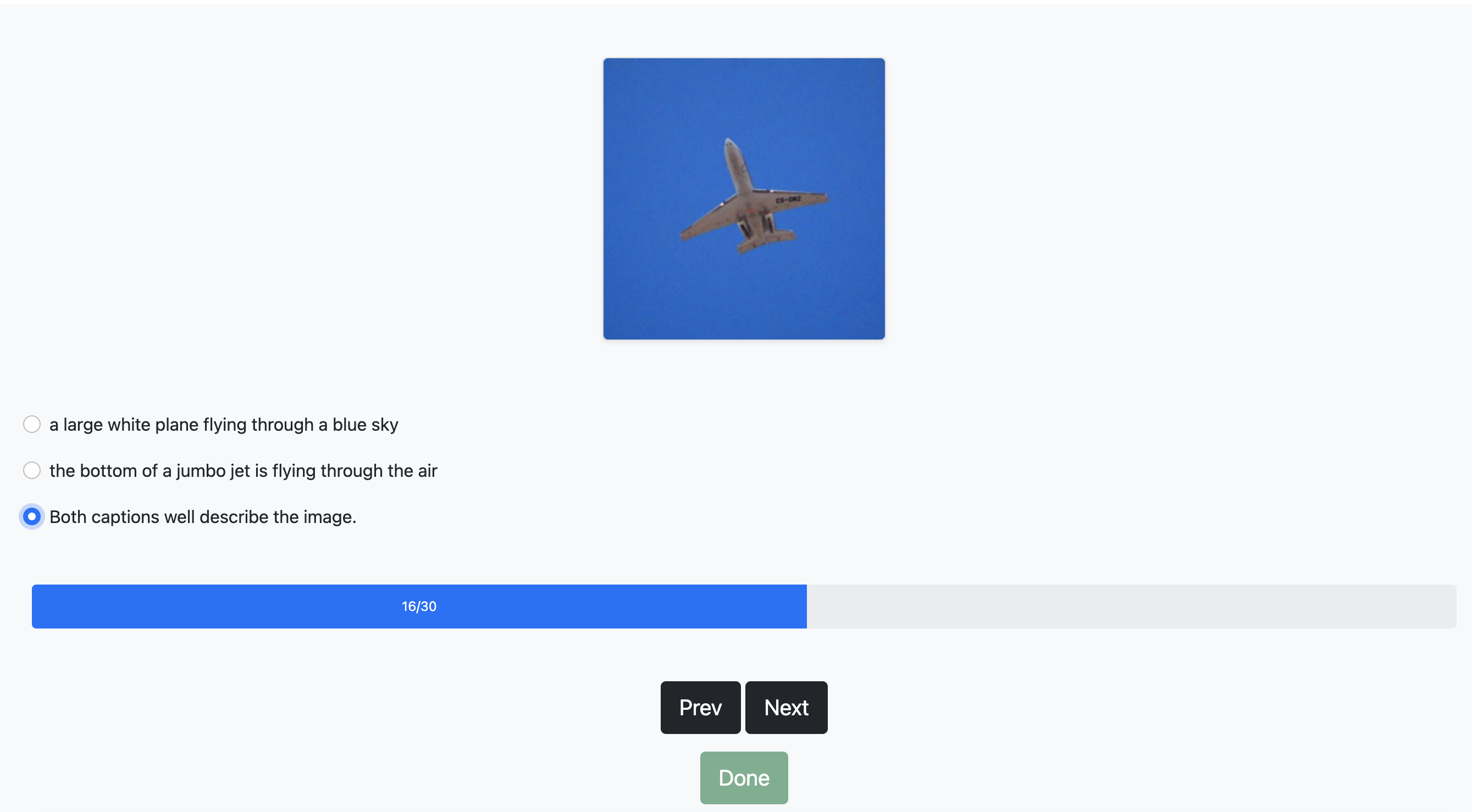}
\caption{Example interface for human evaluation. Random sampled 30 examples are shown to each participant.}
\label{eval_interface}

\end{figure*}

\begin{figure*}[!tb]
  \centering
\includegraphics[width=0.9\linewidth]{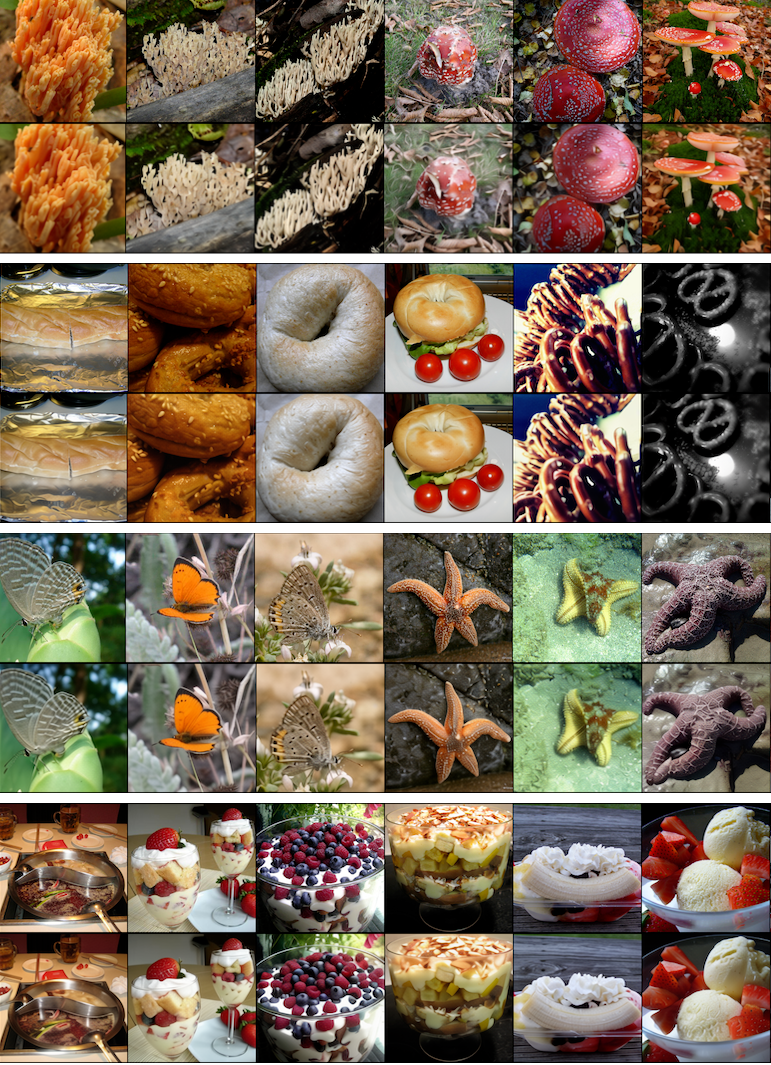}
\caption{
More examples of input images \textit{(top)}
and reconstructions from AugVAE-SL \textit{(bottom)} on Imagenet1K validation set. The resolution of each image is $256 \times 256$ pixels.  }
\label{in_domain}
\end{figure*}

\begin{figure*}[!tb]
  \centering
\includegraphics[width=0.95\linewidth]{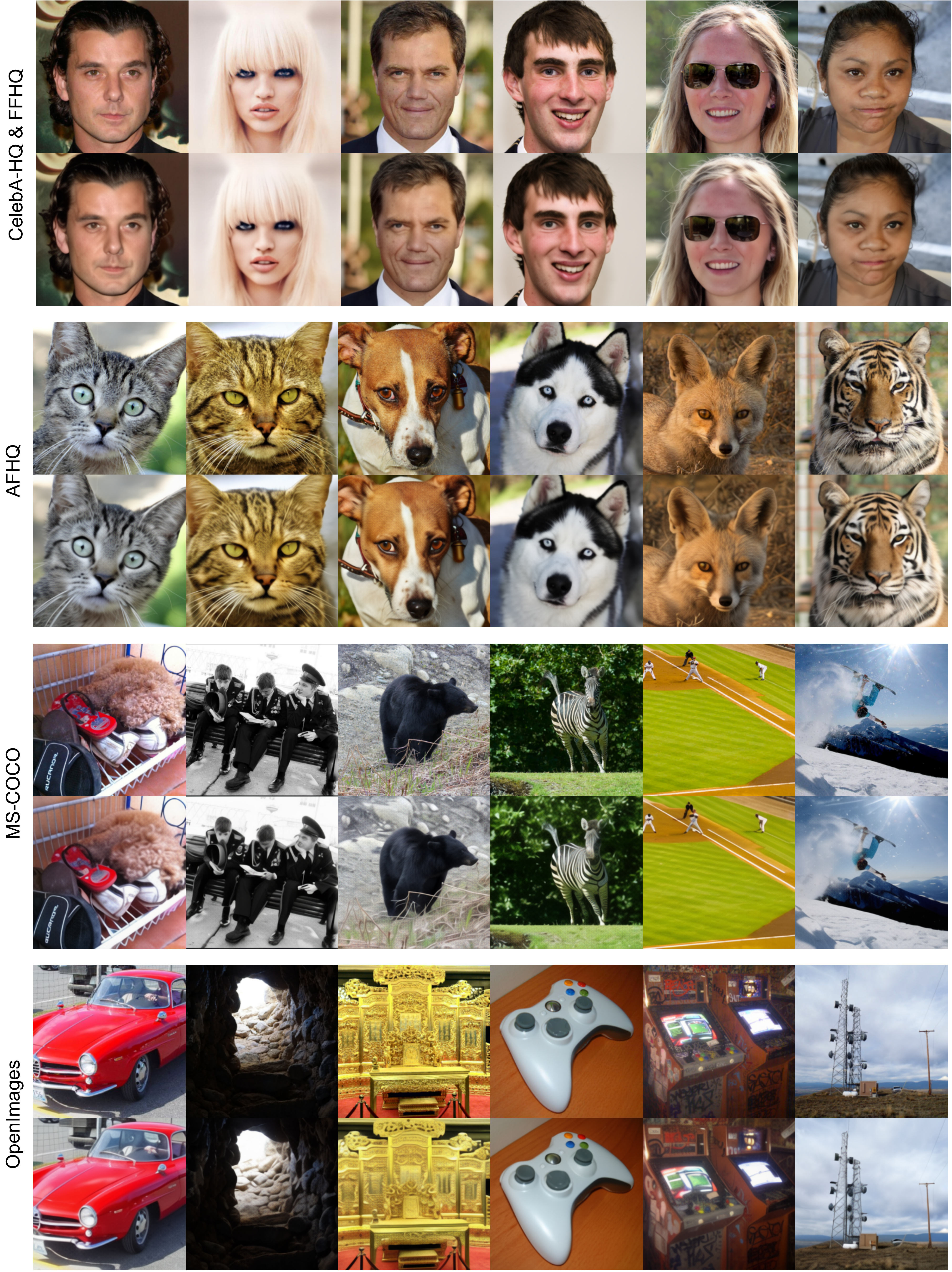}
\caption{
More examples of input images \textit{(top)}
and reconstructions from AugVAE-SL \textit{(bottom)} with unseen image domains ($256 \times 256$ pixels). }
\label{out_domain}
\end{figure*}

\begin{figure*}[!tb]
  \centering
\includegraphics[width=0.9\linewidth]{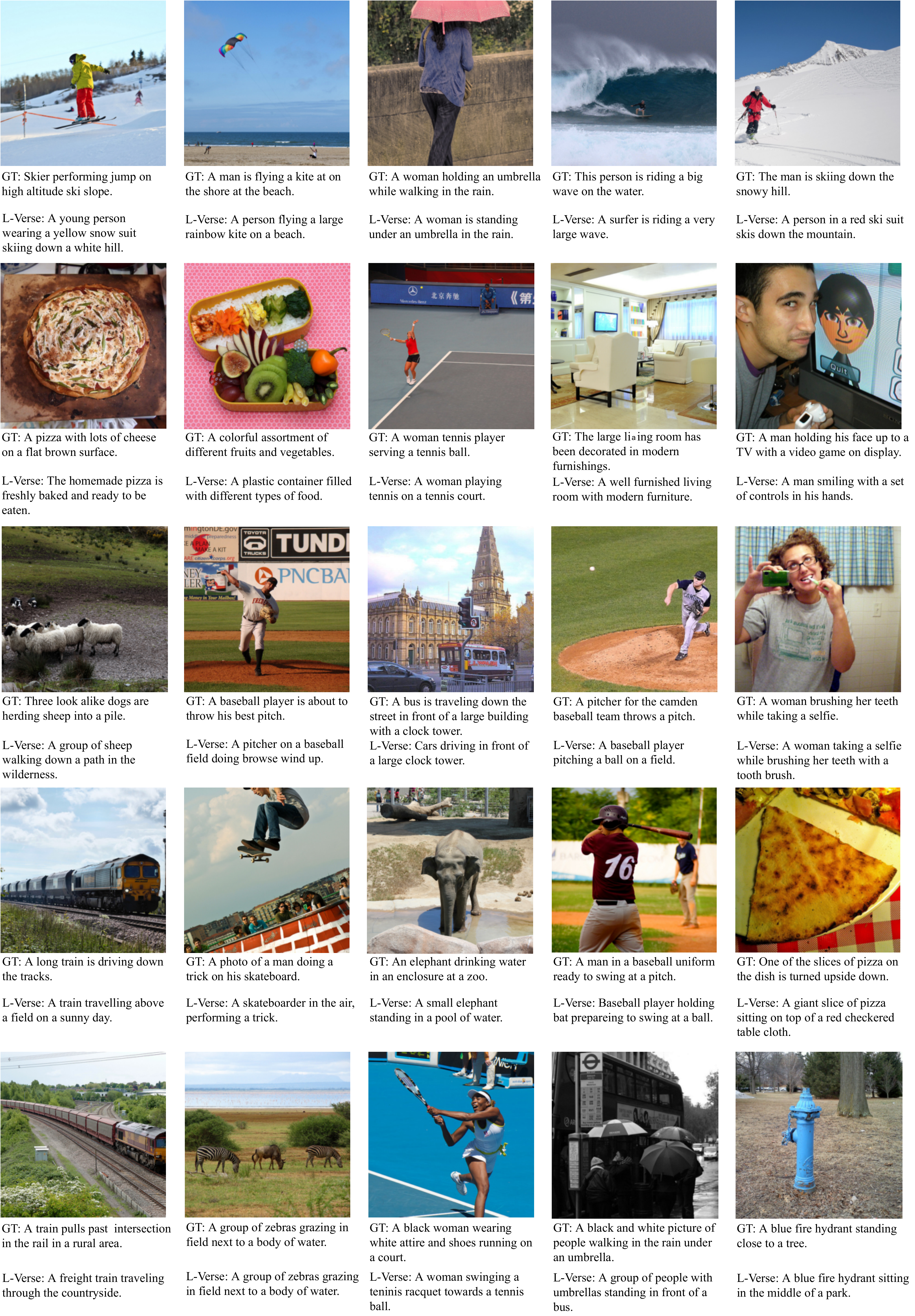}
\caption{
More examples of image-to-text generation on MS-COCO with corresponding ground-truths. }
\label{more_captions}
\end{figure*}

We provide an example task interface of our human evaluation we mentioned in Section \ref{image_text} in Figure \ref{eval_interface}. We also provide more examples of image-to-text generation on MS-COCO Captions in Figure \ref{more_captions}.  All examples in Figure \ref{more_captions} received \textit{``Both captions well describe the image"} in our human evaluation. The resolution of each input image is $256\times256$.

\section{Examples for Text-to-Image Generation}
\label{app_t2i}
\begin{figure*}[!tb]
  \centering
\includegraphics[width=0.9\linewidth]{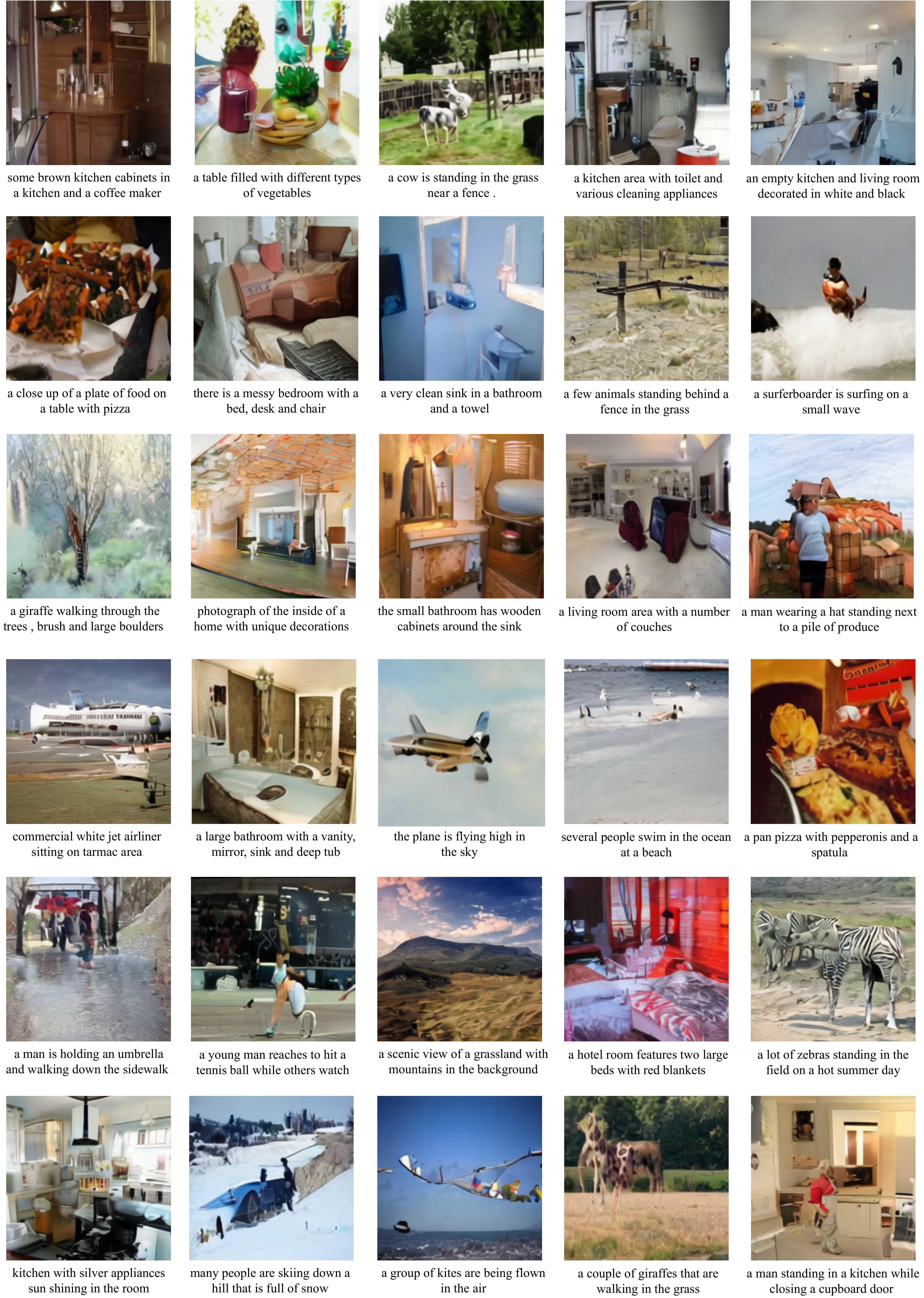}
\caption{
Examples of text-to-image generation on MS-COCO. }
\label{more_images}
\end{figure*}

We provide examples of zero-shot text-to-image generation with L-Verse-CC in Figure \ref{more_images}. Captions are randomly sampled from MS-COCO Captions 2017 validation set. The resolution of each generated image is $256\times256$. 

\section{Discussion}
\label{app_dis}
\paragraph{Bidirectional Learning} L-Verse internally learns a \textit{reversible and densely connected} mapping between images and texts. From this, L-Verse can generate a text or an image in accordance with the given condition without any finetuning or extra object detection framework. Bidirectional learning not only saves time and computational cost for training and application. As we mentioned in Section \ref{training_detail}, our bidirectional approach also mitigates the heterogeneity of data and enables stable \texttt{FP16(O2)} mixed-precision training.

\paragraph {Efficiency} The bidirectional training enables L-Verse to efficiently learn the vision-language cross-modal representation with smaller dataset and model size. L-Verse requires 97.6\% less data (compared to OSCAR \cite{li2020oscar}) for image-to-text and 98.8\% less data (compared to DALL-E \cite{ramesh2021zeroshot}) for text-to-image generation to achieve comparable performances. L-Verse also has 95\% less parameters compared to DALL-E , which makes L-Verse more suitable to the environment with limited computing resources.

\paragraph{Vision-Language Pre-Training} 
Vision-Language (VL) pre-training from OSCAR surely brings positive effects in learning the cross-modal representation. This also follows the current trend of large scale model training: pre-training with a large data set on a general task and fine-tuning with smaller set to solve downstream tasks. Since we mainly focus on the efficiency over the amount of training data and computing resources, VL pre-training is out-of-scope of this work. However, we also believe that combining VL pre-training with bidirectional training will further improve the performance of L-Verse. %We will leave it as a future work.

\paragraph{Large Scale Training}
With limited amount of training data and computational resources, we couldn't consider training L-Verse in larger scale like OSCAR, DALL-E or CogView \cite{ding2021cogview}. Nevertheless, our bidirectionally trained L-Verse shows competitive results to other large scale models.  As 400M well-filtered text-image dataset \cite{schuhmann2021laion400m} has been released recently, we are optimistic about training L-Verse in larger scales.

\paragraph{Zero-Shot Image Captioning}
L-Verse also has an ability to perform zero-shot image captioning when trained on Conceptual Captions (CC) \cite{sharma2018conceptual}. Unlike MS-COCO Captions \cite{lin2015microsoft} which is carefully annotated by humans, images and their raw descriptions in CC are harvested from the web. While texts in CC represent a wider variety of styles, its diversity also adds noise to the caption that L-Verse generates. For this reason, we mainly use L-Verse trained with MS-COCO for the experiment on image captioning.

\paragraph{Potential Negative Impact} Our findings show excellent performance in both image-to-text and text-to-image generation. L-Verse has a wide range of beneficial applications for society, including image captioning, visual question answering, and text visualization. However, there are still potential malicious or unintended uses of L-Verse including image-to-text or text-to-image generation with social bias. To prevent potential negative impact to our society, we provide open access only to AugVAEs for now.

\clearpage



\end{document}